# Building a logical model in the machining domain for CAPP expert systems

**V.V. Kryssanov, A.S. Kleshchev, Y. Fukuda, and K. Konishi**

Recently, extensive efforts have been made on the application of expert system technique to solving the process planning task in the machining domain. This paper introduces a new formal method to design CAPP expert systems. The formal method is applied to provide a contour of the CAPP expert system building technology. Theoretical aspects of the formalism are described and illustrated by an example of know-how analysis. Flexible facilities to utilize multiple knowledge types and multiple planning strategies within one system are provided by the technology.

## 1. Introduction

One of the predicaments of flexible manufacturing is process planning. Although a number of Computer Aided Process Planning (CAPP) systems have been implemented, human planners are still irreplaceable for actual manufacturing. Because process planning requires multiple types of human expertise, there is a common trend to apply knowledge-based techniques for solving the process planning tasks. This circumstance is conducive to developing so-called CAPP Expert Systems (CAPPES).

A few approaches to building CAPPES can be found through means-aids analysis of the research literature since 1980. At the same time, it can be seen that authors' efforts in those papers have mostly been made in special cases of CAPPES implementation, whereas the problem of "How to develop CAPPES" on the whole is still open. Several general conceptions and methodologies for CAPP have been published, but no fairly versatile technology is yet known.

The aim of the paper is to consider the usage of logical models for development of a CAPPES building technology.

### *1.1. Background*

Process planning is a complex activity based on utilizing different kinds of knowledge - declarative as well as procedural - of the machining domain (Wang and Li 1991). To make use of the knowledge in CAPPES, there are three models of knowledge representation often employed:

(production) rules, frames and semantic networks (Kiritsis 1993). The knowledge is usually organized in the knowledge base according to one of the methodological approaches to planning (Barr and Feigenbaum 1981): non-hierarchical, hierarchical, script-based (variant) or opportunistic. It will be noted, that not only the knowledge scheme but, to a considerable degree, the inference mechanisms and the search technique are fixed by the planning methodology chosen.

The earliest CAPPES which appeared in the research literature, such as GARI (Descotte and Latombe 1981), TOM (Matsushima et al. 1982), utilized production rules in its knowledge representation. On the contrary, the frame-based approach was applied to develop the expert systems SIPP (Nau and Chang 1985), HUTCAPP (Mantyla and Opas 1988) and the like. The value of these studies is that they have demonstrated the advantages of dealing with knowledge-based systems for process planning. However, practice has shown that none of the knowledge representation models alone can be sufficient to grasp all the multifarious types of the domain knowledge (Wang and Li 1991). All the CAPPES mentioned above were nothing but demonstration prototypes.

A tendency to use mixed techniques of knowledge representation, including combinations of frames, rules and networks, was inherent to the next generation of CAPPES designed. They are Turbo-CAPP (Wang and Wysk 1987), DPP (Iwata and Fukuda 1989), XROT (Domazet and Lu 1992), etc. As a rule, those systems operated with a restricted piece of the machining domain knowledge and covered a few process planning subtasks, e.g., operations selection, tool selection, grouping operations. The local aims pursued in the systems cause the restricted domain models used and vice versa. The problem of knowledge acquisition for a large and complex domain like the machining domain is still a major bottleneck to the implementation of a considerably full CAPPES (Benjamin et al. 1995).

Another key aspect of the CAPPES development is a planning methodology for generating a process plan, i.e., a method for solving the process planning task. The importance of the appropriate planning strategy choice is proved by extensive investigations on this topic in recent years (e.g., Kruth et al. 1994, Horvath et al. 1996). At present, many of the process planning systems developed employ only one of the planning methodologies for all their various planning activities. These methodologies can be domain independent (DI), which are traditional for AI, as well as domain-dependent (DD), which are specific for CAPP. However, as it was reasonably noted by Mayer et al. (1995), actual process planning is based on a number of different planning manners, and no single strategy alone can be considerably universal for the planning. On the other hand, an attempt to represent more than a few simple subtasks to be solved by different methods in the same CAPPES, leads to the substantially unmanageable domain model with numerous actions, goals, relations and contradictions among them (Opas et al. 1994). There are no technical means (algorithmic, program or hardware based) to subdue this data-set problem, but an adequate formalism is needed.

Thus, the focus of recent research has been on finding flexible facilities to use multiple knowledge types and multiple planning strategies within one system. Few attempts have been reported in the literature to provide a CAPPES development framework capable of facing the realistic needs of CAPP.

An object-oriented approach integrating both DI and DD strategies has been suggested (Yut and Chang 1994). Through interaction of the DI problem-solver with the domain-objects owned domain knowledge, a CAPPES can be adapted to the task. Various unique approaches replacing the problem solver can be applied to the planning, and the specific knowledge as well as the domain-objects can be altered according to the object-oriented paradigm well-known today. Nevertheless, at least two important questions are still uncertain: how to acquire the domain knowledge from experts accurately, and how to solve a planning problem when it is NP-complete (e.g., sequencing of machining operations). The more critical question is whether all the vital domain knowledge can be involved in the planning with this technique.

A cognitive approach has been applied to overcome the problems pointed out above (Mayer et al. 1995). Utilizing metaknowledge extracted from human planning activities (i.e., a process planning ontology) provides the capturing of general properties of the domain. These properties establish the guidelines to acquire the declarative domain knowledge directly from experts and the overall pattern of process planning to apply to manifold planning methods. In fact, Mayer's study introduces a theoretical framework rather than a practical one for constructing an automated process planning system. The questions of verification, validation as well as robustness to metaknowledge shifts for the system on the whole are unsolved within the conception. Moreover, it is not obvious what knowledge representation means is sufficient for all the approach needs, and to what extent the effectiveness of the system depends on the knowledge base size.

Despite the presence of several attempts to solve the problems analogous to those named above (e.g., Horvath et al. 1996), no formal method offered is adequate to stand for the complete life cycle of CAPPES. As a rule, the research results still have a local quality rather than a fundamental one from the technological standpoint though their authors declared more or less global goals. On the other hand, the literature reveals nothing of the systems used in industry, but the only known project having results applicable to actual manufacturing is CAM-I project (Nolen 1989). So, it is hardly possible to generalize successful experiences for a CAPPES building technology.

A universal practice of software development is that to solve a complicated problem the following should be considered (Saiedian 1996). 1) A formal method adequate to design the software (that is "what to do"). 2) A technology based on the method to detail the activity necessary to elaborate the software (that is "how to do"). 3) An application environment of the technology to ensure implementing and maintaining the software (that is "through what to do"). Since a CAPPES is the software for the most part, these issues are crucial for CAPP development. However, this software is of the expert systems' class, and specific requirements must be formulated to address the common problems mentioned. Namely, a formal method for CAPPES design is necessary to embody a knowledge representation means (language) strong enough to specify the domain. A CAPPES building technology should provide the domain analysis including knowledge acquisition from experts as well as validation and verification of the knowledge base. Much of CAPPES' application environment has to be accessible not only to programmers or knowledge engineers but also to domain experts, i.e., human beings beyond the computer sciences. These are only superficial questions, but no attempts to discuss their involvement have been reported in literature.

This paper introduces such a formal method and outlines such a technology.

### *1.2. Glossary*

The following concepts will be used in the paper.

- A *signature* is a finite set of elements called symbols. The symbols can be of three types: objective, functional and predicate ones. A functional symbol as well as a predicate symbol has the arity - an integer more than 0.
- An *algebraic system* of a signature is a couple consisting of a universum and an interpretation. The universum is a set. The interpretation is a mapping of the signature. The interpretation of an objective symbol of the signature is an element of the universum. The interpretation of a functional symbol is a function. The range of definition of the function is a subset of the Cartesian power of the universum (the exponent is equal to the arity of the functional symbol). The range of values of the function is a subset of the universum. The interpretation of a predicate symbol of the signature is a predicate. The range of definition of the predicate is a subset of the Cartesian power of the universum (the exponent is equal to the arity of the predicate symbol).

A fundamental and more formalized view of the above abstractions can be found in (Ehrig and Mahr 1985).

- A *second-order predicate calculus language* is a predicate language where terms and formulas can be in the form: $v(t_1,\ldots,t_n)$; here $v$ is a variable, $t_1,\ldots,t_n$ are terms, $n$ is an arity. The range of values of $v$ is a set of functions (in the case of terms) or predicates (in the case of formulas) with the arity been equal to n.

More details of second-order predicate calculus can be found in (Shapiro 1985). However it is important to notice that in our paper, the second-order predicate calculus language (as well as a first-order one) will be treated as an *abstract language* (McCarthy 1996), i.e., a formal language defined by a collection of abstract synthetic and analytic functions and predicates that form, discriminate, and extract these expressions. This language generally has abstract syntax and admits representations as strings of symbols.

## 2. The theory

Research of the widespread practice of software development shows us that the life cycle of a program can be divided into three principal stages: building the program model, developing the program design, and implementing and maintaining the program. Figure 1 presents the expert system development cycle. The initial point of the process shown in the figure is the creation of an expert system mathematical model. We will consider that the mathematical model of an expert system consists of the domain model underlying this system, the task specification in terms of the domain model (the task is solved by the expert system) and the method for solving the task (the expert system is an implementation of this method). The purpose of this section is to provide the general ideas underlying the creation of a CAPPES model with particular emphasis on building the domain model as the most important part of CAPP. Other aspects of the theory can be found in (Kleshchev 1994), (Artemjeva et al. 1994) and (Artemjeva et al. 1995).

## *2.1. Logical models*

Let us define a knowledge representation means capable of representing the domain knowledge. This means, called a class of logical models, has a declarative semantic, i.e., this semantic specifies correspondence between knowledge and reality.

A set $\Sigma = \Sigma_0 \bigcup \Sigma_1 \bigcup \ldots \bigcup \Sigma_n$, $n > 0$ will be called a signature of *n*-th order if $\Sigma_i \bigcap \Sigma_j = \varnothing$ for $i \neq j$, and $\Sigma_i$ is a signature for $i = 0, 1, \ldots, n$. We define a logical model $\Omega$ with the signature $\Sigma$ of *n*-th order as a tuple $\Omega^n = \langle \Phi, \mathrm{A} \rangle$. Here $\mathrm{A} = \{\mathsf{A}_0, \mathsf{A}_2, \ldots \mathsf{A}_n\}$ if $n > 1$, and $\mathrm{A} = \{\mathsf{A}_0\}$ if $n = 1$; $\mathsf{A}_0$ is an algebraic system of the signature $\Sigma_0$; for $i = 2, \ldots, n$ $\mathsf{A}_i$ is an algebraic system of the signature $\Sigma_i$. $\Phi$ is a finite set of quantifier-free logical formulas written by a predicate calculus language in terms of the signature $\Sigma$ (that stipulates applying predicate calculus logic rather than a programming language). The language is a first-order language if $n = 1$ and a second-order language if $n > 1$.

The objective symbols of $\Sigma_0$ will be named constants, the functional symbols of $\Sigma_0$ will be named signs of operations and functions, the predicate symbols of $\Sigma_0$ will be named signs of relations. The symbols of $\Sigma_1$ will be named objective, functional and predicate unknowns. For $i > 1$ the symbols of the signature $\Sigma_i$ will be called names of parameters (objective, functional and predicative) of the $(i-1)$-th order. We will consider the interpretations of all the functional and predicate symbols of $\Sigma_0$ as calculable functions and predicates. The universum of the algebraic system $\mathsf{A}_2$ consists of the objective symbols of the signature $\Sigma_0$ and all the symbols of the signature $\Sigma_1$. For $i = 3, \ldots, n$ the universum of $\mathsf{A}_i$ is the union of the universum of $\mathsf{A}_{i-1}$ and the signature $\Sigma_{i-1}$. For $i > 1$ the interpretations of all the functional and predicate symbols of the $\Sigma_i$ are functions and predicates with finite ranges of definition. So, all these functions and predicates can be represented by finite tables.

Every variable being a member of a formula of $\Phi$ has an order, i.e., an integer between 1 and *n*, and a range of values. If the order of a variable is equal to 1 then its range of values is a subset of the universum of $\mathsf{A}_0$. For a variable of the *i*-th order, $i > 1$, the range of values is a subset of the universum of $\mathsf{A}_i$.

An algebraic system $\mathsf{A}$ of the signature $\Sigma_1$ will be called relevant to the logical model $\Omega^n$ if the universum of $\mathsf{A}$ is a finite subset of the universum of $\mathsf{A}_0$. A substitution $\lambda = \{v_1/a_1,\ldots,v_k/a_k\}$ is relevant for a formula $\phi \in \Phi$ and an algebraic system $\mathsf{A}$ if all the variables being members of $\phi$ are included in the set $\{v_1,\ldots,v_k\}$ and if for any $j = 1,\ldots,k$ the value $a_j$ belongs either to the intersection of the range of values of $v_j$ and the universum of $\mathsf{A}$ ($v_j$ is a variable of the first order) or to the range of values of $v_j$ (the order of $v_j$ is more than 1). A formula $\phi \in \Phi$ and an algebraic system $\mathsf{A}$ are in agreement if $\phi$ is true for any relevant substitution, when all the symbols of $\Sigma_0$ are interpreted in the algebraic system $\mathsf{A}_0$, all the symbols of $\Sigma_1$ are interpreted in the algebraic system $\mathsf{A}$, for $i > 1$ all the symbols of $\Sigma_i$ are interpreted in $\mathsf{A}_i$. The set of all the relevant substitutions is a finite one. A relevant algebraic system $\mathsf{A}$ is a solution for $\Omega^n$ if every formula entering $\Phi$ and the algebraic system $\mathsf{A}$ is in agreement.

A logical model of a signature $\Sigma_0 \cup \Sigma_1 \cup \Sigma_2' \cup \ldots \cup \Sigma_m'$ and a logical model of a signature $\Sigma_0 \cup \Sigma_1 \cup \Sigma_2'' \cup \ldots \cup \Sigma_n''$ will be taken for equivalent systems if they have the same solution set. One can prove the theorem about decreasing the order of a logical model (Artemjeva et al. 1996): for any logical model $\Omega^n$, $n > 1$, there is the logical model $\Omega^{n-1}$ being equivalent to $\Omega^n$. This theorem allows us to replace a logical model of a higher order by the equivalent logical model of the first order.

## *2.2. Domain logical models*

In this section, we will show how to use the logical models $\Omega^n$ defined above to model the domain.

We will assume that a domain is characterized by a professional activity. This activity consists of solving different tasks. Task solving requires professional knowledge, the same for all the tasks. The professional knowledge, and the input data and the output data for every task can be represented verbally. We will regard the set of domain objects, both physical and conceptual, the reality and the knowledge as the main domain components. The information about the domain objects is used as the input, output and intermediate data of the task of the professional activity. The professional activity takes place in the domain reality. The professional knowledge is the basis of the activity.

We will consider that every domain object can belong to a magnitude. A magnitude is a set of domain objects having properties common to all the objects of this magnitude. A domain also includes a set of scales. Each scale corresponds to a magnitude and is its mathematical model representing all the common properties of magnitude objects. Every scale has the set of objective, operation and relation symbols for representing the common properties of magnitude objects. These symbols form the signature $\Sigma_0$. They have an interpretation, generally accepted or special. This interpretation is simulated by the algebraic system $\mathsf{A}_0$. The professional activity uses scales and scale values instead of magnitudes and magnitude objects. For every domain we will call the set of domain scales the domain scale system. The domain scale system allows us to represent mathematically the properties of the domain objects in the domain model.

Let a domain logical model be a logical model $\Omega^n$ consisting of the domain scale system formalized as $\mathsf{A}_0$, the reality model - as a solution set for $\Omega^n$, and the knowledge model - as a set of facts represented by the algebraic systems $\mathsf{A}_2,\ldots,\mathsf{A}_n$ and a set of agreements about the domain representation written as the set of formulas $\Phi$.

We will regard the reality as an infinite set of separated situations. Each situation is represented by information related to a task in the domain. This information affects a finite set of domain objects represented by scale values. Every domain object belonging to the set will be called a situation object.

The domain reality has the structure consisting of finite sets of roles, notions and relations among domain objects. Any of these sets may be empty but not all. Every role defines for every situation an object of this situation. Every notion defines for every situation a set of objects of this situation. Every relation defines for every situation a set of $k$-tuples, $k > 1$, consisting of objects of this situation.

The objective symbols of the signature $\Sigma_1$ are the designations of the reality roles. The functional symbols of $\Sigma_1$ are the designations of the functional relations of the reality. The predicate symbols of $\Sigma_1$ are the designations of the unfunctional relations of the reality. There are the monadic predicate symbols of $\Sigma_1$, which are distinguished as a special case and are the designations of the reality notions.

The logical model is a domain logical model if a model of any domain situation is an algebraic system $\mathsf{A}$ relevant to the logical model. So, the universum of $\mathsf{A}$ is a model of the situation object set, and the interpretation of the signature $\Sigma_1$ in $\mathsf{A}$ is a model of the situation structure. The same domain can have a few models. Logical models of different orders may be among them.

At first, let us consider the case where a logical model of the first order is a model of the domain. Then the domain knowledge can be conceived as a finite set of statements about properties of all the situations of the reality. A set of logical formulas $\Phi$ is the domain knowledge model so that every statement of the domain knowledge has a formula $\phi \in \Phi$ as its model. If each $\phi \in \Phi$ is an implication formula then $\Phi$ is a declarative analog of a production system. Otherwise $\Phi$ can be considered as a declarative analog of a constraint system.

For a domain logical model of the second order, the objective symbols of $\Sigma_0$ and all the symbols of $\Sigma_1$ will be called as terms of the first order. The domain knowledge is considered as a finite set of facts and a finite set of agreements about the reality and knowledge representation. A fact is represented by a tuple consisting of terms of the first order. The facts with a similar sense represented by $m$-tuples ($m > 0$) form a relation among terms of the first order. The functional and predicate symbols of $\Sigma_2$ are the designations of the relations. If $m = 1$, then such a relation will be called a domain metanotion. The universum of $\mathsf{A}_2$ is the set of terms of the first order. The interpretation of the symbols of $\Sigma_2$ in $\mathsf{A}_2$ is a model of the relations among the terms of the first order. It can be regarded as a declarative analog of a semantic hypernetwork. The set of agreements about the reality and knowledge representation is a finite set of statements (metaknowledge) governing interconnection between the knowledge and reality, and $\Phi$ is a model of this set.

For the case of a domain logical model of the third order, the set of relations among terms of the first order is the same as for a model of the second order. All the terms of the first order and all the symbols of $\Sigma_2$ will be called in the same way as terms of the second order. The set of agreements about the reality and knowledge representation is represented by a finite set of relations among terms of the second order and a finite set of agreements about the representation of the agreements about the reality and knowledge representation. The functional and predicate symbols of $\Sigma_3$ are the designations of the relations among terms of the second order. The universum of $A_3$ is the set of terms of the second order. The interpretation of the symbols of $\Sigma_3$ in $A_3$ is a model of the relations among the terms of the second order. $\Phi$ is a model of the set of agreements showing representation of the agreements about the reality and knowledge representation (metametaknowledge). The interpretation of the symbols of $\Sigma_2$ in $A_2$ and the interpretation of the symbols of $\Sigma_3$ in $A_3$ can be considered as two parts of one semantic hypernetwork. The case of a domain logical model of the order more than the third can be similarly considered.

The set of formulas $\Phi$ generally consists of the integrity restrictions of all situation models (this subset may be empty), the integrity restrictions of all facts in the knowledge model, and the formulas determining the connection between the set of all facts in the knowledge model and the models of situations. Every formula in $\Phi$ must be true for all relevant substitutions, instead of all variables being members of this formula. This condition for the connection between the knowledge model and the reality model permits us to define the scale system and the knowledge model of the domain logical model only and to get the reality model from the condition for the connection. (As can be seen, the latest proposition is just a generalization of the software development process.)

Each of the algebraic systems $A_i$, $i > 1$ in the knowledge model represents a finite set of the facts on its own level of abstraction. We will call every functional (predicative) symbol of $\Sigma_i$ as a name of a functional (unfunctional) relation of *i*-th order. Every fact is a tuple of such a relation.

We will suppose that a logical model is an adequate model of the domain if every situation of the reality has a model in its solution set, and every solution is an adequate (i.e., enough for what is required) model of a situation in reality. Since the set of situations in reality is usually infinite, in practice only a weaker criterion can be used: a logical model is not an adequate model of the domain if such a situation exists in reality in a way that its model does not belong to the solution set.

*2.3. Task specification form*

We will consider a specification of any task for a logical model $\Omega^n$ of the signature $\Sigma$ as a 4-tuple $\mathrm{T} = \langle \Omega^n, \Delta, \Psi, \Pi \rangle$. Here $\Delta$ is a representation of the input data and conditions of the task. $\Delta$ is a set of quantifier-free formulas written by the first-order predicate calculus language in terms of $\Sigma_0 \bigcup \Sigma_1$. Every formula in this set must include at least one symbol of $\Sigma_1$. $\Psi$ is a representation of the optimization criterion of the task and is a predicate defined on the solution set of $\Omega^n$. If the task has no optimization criterion then we will assume $\Psi \equiv true$. $\Pi$ is a representation of the output data of the task in terms of $\Sigma_1$.

To define a task solution, we will build a new logical model $\Omega_t^n$ of the signature $\Sigma$: $\Omega_t^n = \langle \Phi_t, \mathrm{A} \rangle$, where $\Phi_t = \Phi \bigcup \Delta$. The solution set of $\Omega_t^n$ is a subset of the solution set of $\Omega^n$. Let $S_t$ be a designation of the set consisting of such solutions of $\Omega_t^n$ for which $\Psi = true$. If $S_t$ is empty then the task has no solutions. Otherwise, if $\mathsf{A} \in S_t$ then a task solution is the following mapping. For any $\pi \in \Pi$ the result of mapping $\pi$ is the interpretation of $\pi$ in $\mathsf{A}$.

## 3. Technological interpretation of the theory

The conception previously described is the foundation of the CAPPES building technology developed in the research. Figure 2 illustrates the principal components of the technology.

The starting point of the technology is the creation of the CAPPES mathematical model. On this stage, an analysis of the domain is executed by experts in machining and knowledge engineers together. The analysis procedure is initiated by finding the scale system practiced in the domain. The sets of the scale values for all the scales and representation of the values on the scales must be determined. All the operations and relations concerned with the scales should be identified and strictly described. A unified representation of the domain scale system is the algebraic system $\mathsf{A}_0$.

- *Examples.* Finding the scale system for the machining domain might be considered as a part of the activity already performed in the framework of the development of a standard like ISO 10303 (STEP) for representation of whole product model data (ISO 1994). It was found that in the machining domain the scale system consists of such subsystems as the dimensional scale system (for representation of numerical physical magnitudes), the scalar scale system (for representation of objects having names), the structural scale system (for representation of objects having an internal structure) and also the scale of 3-dimensional solids (for representation of part projects after CAD systems), the scale of parts (for representation of parts), the geometrical feature scale system (for representation of geometrical features) and the technological feature scale system (for representation of technological features).

A practical instance of the structural scales is the scales obtained by the authors through analysis of know-how collected from manufacturing companies in Japan (TRI 1996). Some fragments of these scales are the following (in terms of the originals):

*workpiece material =* *structure* *(classification: scalar scale; property: scalar scale; hardness: HB)*

(a workpiece material has such parameters as classification and property on their own scalar scales, e.g., {alloy steel, cast iron, …}, {SS41, S50C, …} and hardness on the HB scale);

*end-mill cutting operation =* *structure* *(... tool: end-mill; cutting depth: mm; cutting width: mm; cutting speed: m/min; feed: mm/min; ... ; cutting fluid: scalar scale; ...)*

(the explanation can be analogously made);

*end-mill =* *structure* *(type of end-mill: scalar scale; diameter: mm; ... ; cutting tooth length: mm; ... ; number of cutting teeth: integer; ...)*

All the terms being members in these definitions belong to the signature $\Sigma_0$.

The next stage of the domain analysis is representation of domain situations. All the necessary situations of the domain must be discerned and grouped into classes corresponding to classes of the tasks solved. All the roles, notions and relations among the domain objects of the situations - the structure of the domain reality - should be found, and designations for them should be introduced. These designations constitute the signature of the reality model $\Sigma_1$. Besides, it is necessary to be sure that for all the situations of all the classes their adequate models can be built via $\Sigma_0 \cup \Sigma_1$. It can be seen that this activity is of finding the domain ontologies, and the symbols of $\Sigma_1$ are the formal specification elements, with which one represents the ontologies.

- *Examples.* A situation concerning calculating the cutting parameters of an NC-operation with end-mill is presented in Figure 3. The situation found in the know-how collection is an example of the local subtask of the machining parameters determination subtask of process planning. Situations like this can be grouped into the class where workpiece material is defined as an example of the role. We can introduce $x \in \Sigma_1$ as the designation of an unknown belonging to the workpiece material scale (*x* is the workpiece material in a local subtask of the situation class). We can also presuppose that such situation objects as *Feed* and *Cutting speed* are referred to the situation notion *Standard cutting parameters for end-mill.* (This notion may pertain to the metanotion *Cutting parameters for end-mill.*)

Further, it is necessary to decide between the orders of the domain logical model to be built. Evidently, if the domain logical model order is equal to one then the domain knowledge model is a set of formulas $\Phi$ only. In this case, there is no way to view the whole set of these formulas for experts to be able to choose the formulas belonging to $\Phi$. On the contrary, if the model order is more than one then the domain knowledge model includes a set of facts. In this case, there is a way to view the whole set of these facts for experts to be able to choose the facts belonging to the knowledge model. Managing a set of facts (especially editing this set) is considerably easier than managing a set of formulas. Moreover, to work out a task solving method, it is necessary to take into account only the set of formulas rather than the set of facts. One should go from the domain logical model of a lower order to the model of a higher order if $\Phi$ consists of many formulas. It permits us to decrease the amount of formulas. Hereafter, we will discuss the case of the domain logical model of the order more than one, as preferable from the technological point of view.

Finally, the knowledge representation should be completed. It is necessary to identify all the relations among terms of the first order and assign designations for them - terms of the second order. These designations form the signature of the knowledge model $\Sigma_2$. (If a model of an order *n*, $n > 2$ is built, this procedure must be consecutively repeated for terms of *m*-th order, $m = 1, \ldots, n-1$. Then, the signature of the knowledge model is $\Sigma_2 \cup \ldots \cup \Sigma_n$). All the relations among the domain terms eventuate the knowledge base scheme that can be thought of as a domain ontology.

- *Examples*. The following functional relations were found among the terms of the situation shown in Figure 3:

  *b1* - connecting classification, property (scalar scales) and hardness (HB) of workpiece material with cutting speed (interval of m/min)*;*

  *b2* - connecting diameter, cutting depth and cutting width (mm) with feed factor*;*

  *b3* - connecting classification, property (scalar scales) and hardness (HB) of workpiece material with work material factor*;*

  *b4* - connecting type of end mill (scalar scale), cutting tooth length (mm) and number of cutting teeth with end mill factor*;*

  *b5* - connecting cutting fluid (scalar scale) with cutting fluid factor.

  The symbols *b1*, ... , *b5* belong to the signature $\Sigma_2$.

According to the representation of the situations and the knowledge representation made, agreements about reality and knowledge representations (i.e., a conceptualization of the connection between the reality and the knowledge) are formulated by knowledge engineers. These agreements organize the set of knowledge base formulas $\Phi$ and must be written with the second-order predicate calculus language in terms of the signature $\Sigma$.

- *Examples*. For the situation shown in Figure 3 the following logical formulas can be written in the terms introduced above:

$$\begin{aligned} &\mathit{cutting\ speed}.v1 \in b1(\mathit{classification}.x, \\ &\qquad \mathit{property}.x, \mathit{hardness}.x); \end{aligned} \tag{1}$$

$$\begin{aligned} \mathit{feed}.v1 = {} & b2(\mathit{diameter}.\mathit{tool}.v1, \mathit{cutting\ depth}.v1, \\ & \mathit{cutting\ width}.v1) \times b3(\mathit{classification}.x, \\ & \mathit{property}.x, \mathit{hardness}.x) \times \\ & \times b4(\mathit{type\ of\ end\ mill}.\mathit{tool}.v1, \\ & \mathit{cutting\ tooth\ length}.\mathit{tool}.v1, \\ & \mathit{number\ of\ cutting\ teeth}.\mathit{tool}.v1) \times \\ & \times b5(\mathit{cutting\ fluid}.v1) \times 10000 \times \\ & \times \mathit{cutting\ speed}.v1/\pi/\mathit{diameter}.\mathit{tool}.v1; \end{aligned} \tag{2}$$

where *v1* is a variable designating an object of the *end mill cutting operation* scale.

On the basis of the knowledge representation completed, a set of facts for each of the relations among domain terms should be formed. For this purpose, an Interactive Semantic Editor (ISE) of the knowledge base can be developed using the knowledge base scheme (the domain ontology obtained) as a foundation (Kleshchev 1994). The ISE works as the main instrument to acquire the set of facts (algebraic systems $A_2,\ldots,A_n$) from experts in a form that is natural to the domain. (It can be seen, that such activities would result in a domain knowledge fund.) If the set of facts exists, model validation for the criterion of adequacy should be performed, whereupon building the domain logical model is completed. Figure 4 summarizes a pattern of the machining domain logical model having a solution set as the set of process plan models. Later, the set of facts and the knowledge base scheme are used by a system for automatic generation of the tests for CAPPES. This system can execute preliminary verification and validation of the knowledge base and database.

The next step of the technology is specification of the task(s) that is to be solved with the expert system. This procedure is carried out by experts and knowledge engineers together according to the task specification form (see the theory).

Methods for the complex process planning should be worked out by knowledge engineers. Different reasoning domain-dependent approaches can be used within one CAPPES depending on the type of planning problem. In some cases pure mathematical subtasks can be extracted. They are to be solved by domain-independent mathematical methods to get sufficiently high efficiency. All the methods must be formalized and represented by using the domain logical model built as the basis.

It should be noted, that the well-known practice of CAPPES development was that the task solved with the expert system was determined by the reasoning method rather than by a formal specification of the task. Therefore, the question about a correspondence between the set of task solutions and the set of inference results was meaningless. Working out a formal specification of the task independently from a reasoning method makes this question meaningful. Studying reasoning methods consists in proving theorems like this: the set of inference results is equal to the set of task solutions. Another direction of studying reasoning methods could be evaluating the complexity of these methods.

Since some of the process planning tasks are NP-complete, methods of their solving can be effective only if one takes into account a specificity of the domain knowledge. This specificity is represented with the set $\Phi$ in the knowledge model. Besides, the set $\Phi$ can play the role of restrictions for parameters connected. The suggested approach allows us to use the restrictions as specifications of task solving methods to be worked out. Such ideas are traditional mathematics.

To improve a reasoning method means to modify it in such a way as to decrease its complexity. There are three main sources for that. The first is adaptation of the method to particular properties of domain logical models. The second is adaptation of the method to particular properties of the task input data. The third is transformation of domain logical models of a high order to those of the first order on the basis of the theorem about decreasing order as follows. If the task is solved for a domain logical model of a high order then data processed with the reasoning method consist of input data of the task and the set of facts of the knowledge model. Since the set of facts includes an extremely large number of facts, then inference complexity is extremely high. It has been found experimentally that going from a model of a high order to the model of the first order significantly decreases complexity of inference. Such transformations can be executed by the knowledge base translator.

The approach suggested here permits us to analyze the different problem solving methodologies which evolved and to select the most-efficient ones for CAPP. Such investigations might be accomplished in the framework of a project like the REAL_WORLD project (Tipnis 1996) directly if a domain logical model has been built.

## 4. Conclusions

Research carried out in a number of countries during the last twenty years has shown that the classical techniques of developing expert systems do not give us a way to create a complete CAPPES. Therefore, it is necessary to look for distinctly new approaches to solve the CAPP problem. In this study, a new formal method was introduced and applied to give a view of the contour of a CAPPES building technology.

The general class of declarative logical models for modeling the domain was described. This class is sufficiently general to include analogues of the classical knowledge representation models.

The criterion of adequacy of a model of the class to the domain was formulated. The formal framework to specify the planning tasks was suggested. Further, the technological interpretation of the theory was described. This interpretation constitutes an outline of an application environment of the CAPPES building technology. The pattern of a machining domain logical model was proposed. Using a simple example, it was shown how to analyze the domain knowledge presented by know-how to build a part of the domain knowledge model.

**Acknowledgments**

This research has been supported by the Russian Fundamental Research Foundation (grants no. 93-013-17377, 96-01-00175) and the New Energy and Industrial Technology Development Organization, Japan (the Joint International Project for Development of the Production Design Technology for Machining). It is a pleasure to acknowledge the continued contributions of Dr. Masaru Sugiura and Academician Veniamin P. Myasnikov in the research.

Figure 1. Expert system life cycle.

Figure 2. Schematic diagram of the CAPPES building technology.

Figure 3. An example of the local subtask of the machining parameters determination subtask of process planning.

Figure 4. Scheme of a logical model of the machining domain.

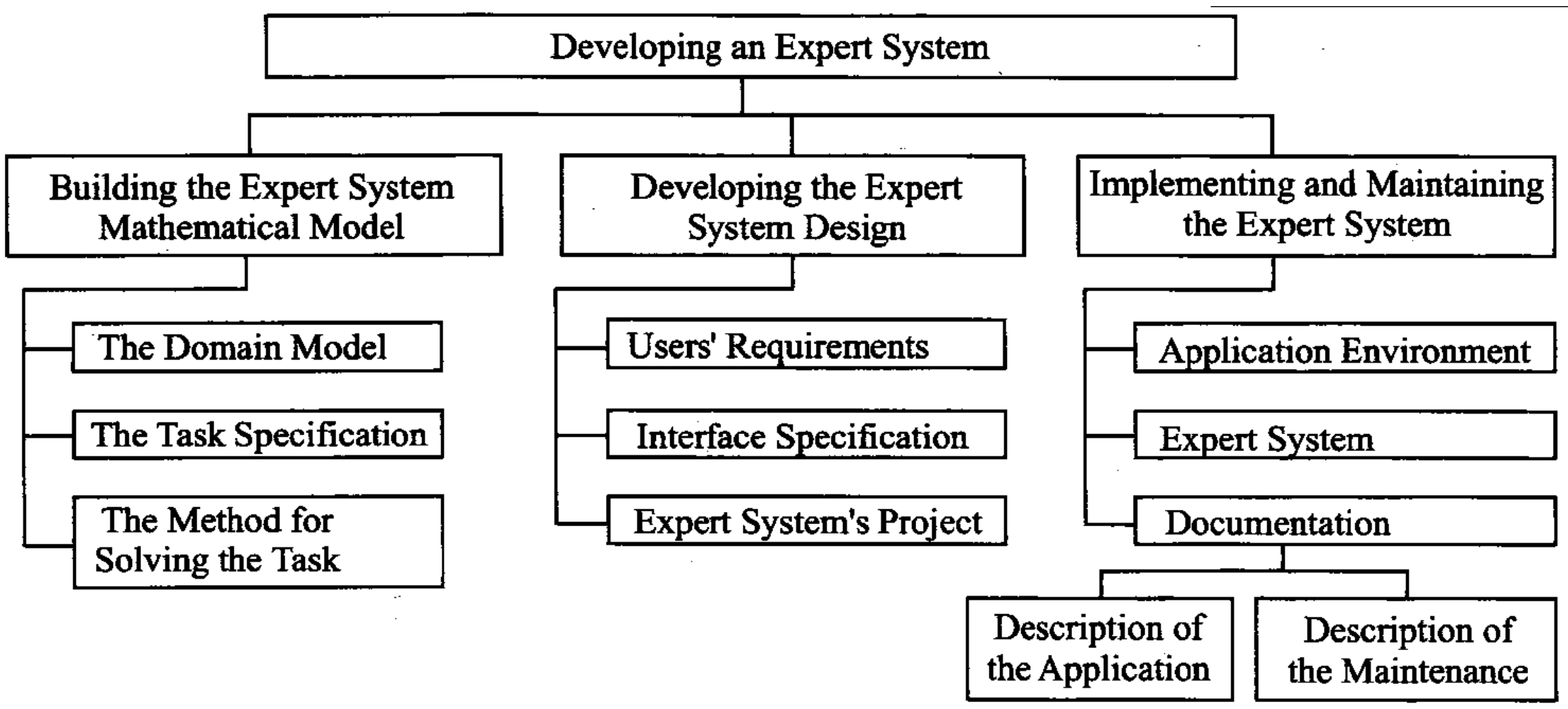
Developing an Expert System
Building the Expert System Mathematical Model
Developing the Expert System Design
Implementing and Maintaining the Expert System
The Domain Model
The Task Specification
The Method for Solving the Task
Users' Requirements
Interface Specification
Expert System's Project
Application Environment
Expert System
Documentation
Description of the Application
Description of the Maintenance

Figure 1.

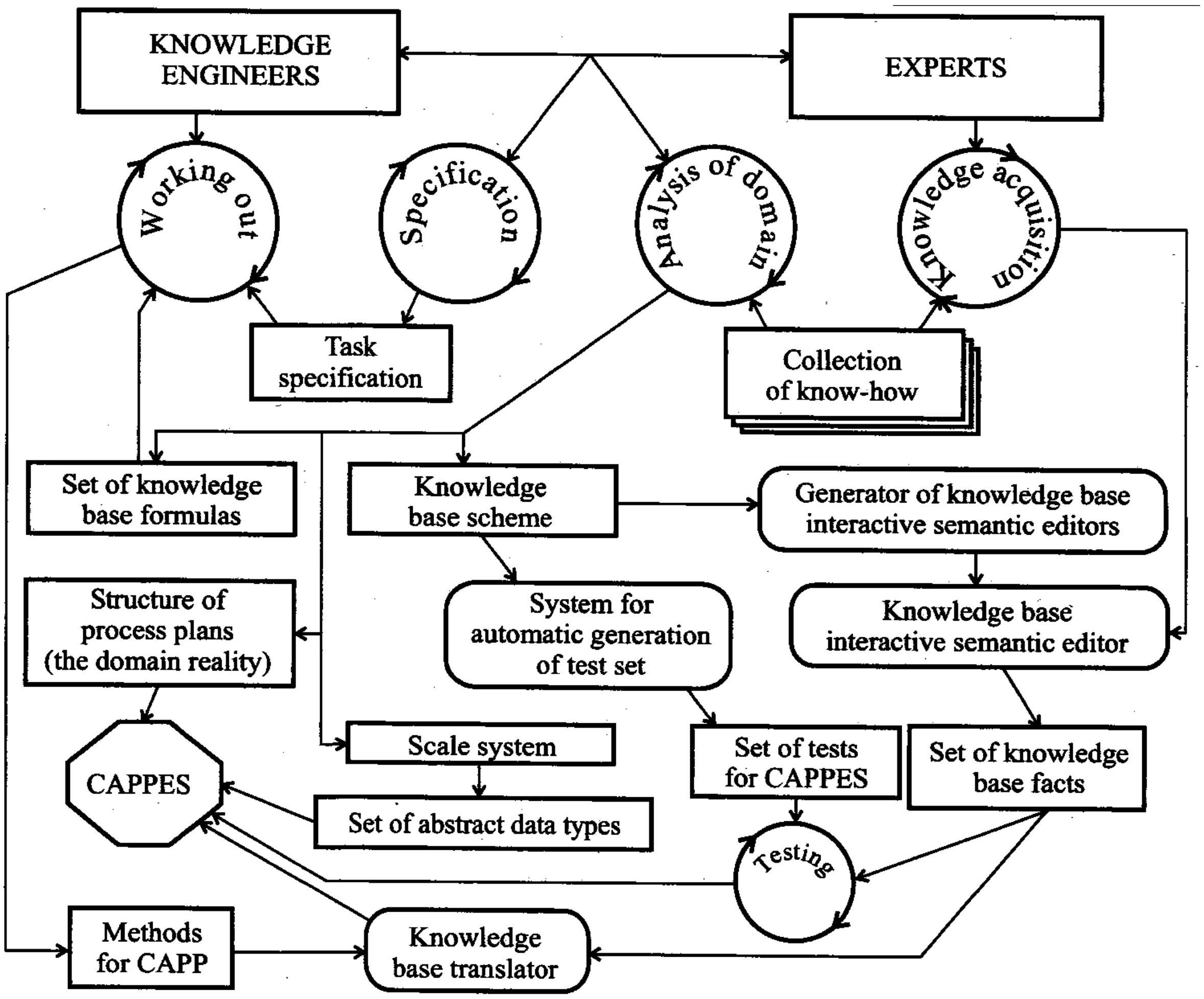
KNOWLEDGE ENGINEERS
EXPERTS
Working out
Specification
Analysis of domain
Knowledge acquisition
Task specification
Collection of know-how
Set of knowledge base formulas
Knowledge base scheme
Generator of knowledge base interactive semantic editors
Structure of process plans (the domain reality)
System for automatic generation of test set
Knowledge base interactive semantic editor
CAPPES
Scale system
Set of tests for CAPPES
Set of knowledge base facts
Set of abstract data types
Testing
Methods for CAPP
Knowledge base translator

Figure 2.

**Feed Factor**

| D \ Ad \ Rd | | 0.1 mm | 0.3 | 0.5 | 1.0 |
|---|---|---|---|---|---|
| Ø1 | 0.5 mm | ... | ... | ... | ... |
| | 1.0 | ... | ... | ... | ... |
| | 1.5 | ... | ... | ... | ... |
| | 2.0 | ... | ... | ... | ... |
| | | | | | |

Note) 1. Ad and Rd are actual dimensions (in mm).
2. Cutting depth Rd equal to the end mill diameter (D) assumes grooving.

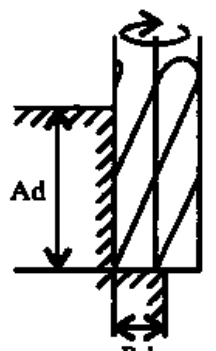

**Work Material Factor**

| Classification | Property | Hardness | Cutting speed m/min | Work material factor |
|---|---|---|---|---|
| Alloy steel | SS41 | HB118 | ... | ... |
| | S50C | HB202 | ... | ... |
| | SK7 | HB210 | ... | ... |
| | SNC3 | HB295 | ... | ... |
| | | | | |

Note) 1. The value of cutting speed is that in the case where nonaqueous cutting fluid is used.
...

$$\text{Feed (mm/min)} = \text{Feed Factor} \times \text{Work Material Factor} \times \text{End Mill Factor} \times \text{Cutting Fluid Factor} \times \frac{10000 \times \text{Cutting speed}}{\pi \times D}$$

**End Mill Factor**

| Type \ Edge length \ Number of edges | | 2 | 4 | 6 |
|---|---|---|---|---|
| Square end mill | Short ... *mm* | ... | ... | ... |
| | Standard ... *mm* | ... | ... | ... |
| | Long ... *mm* | ... | ... | ... |
| | For grooving ... *mm* | ... | ... | ... |
| | | | | |

**Cutting Fluid Factor**

| Type of cutting fluid | Cutting fluid factor |
|---|---|
| Nonaqueous cutting fluid | ... |
| Dry cutting | ... |
| Aqueous cutting fluid | ... |

Figure 3.

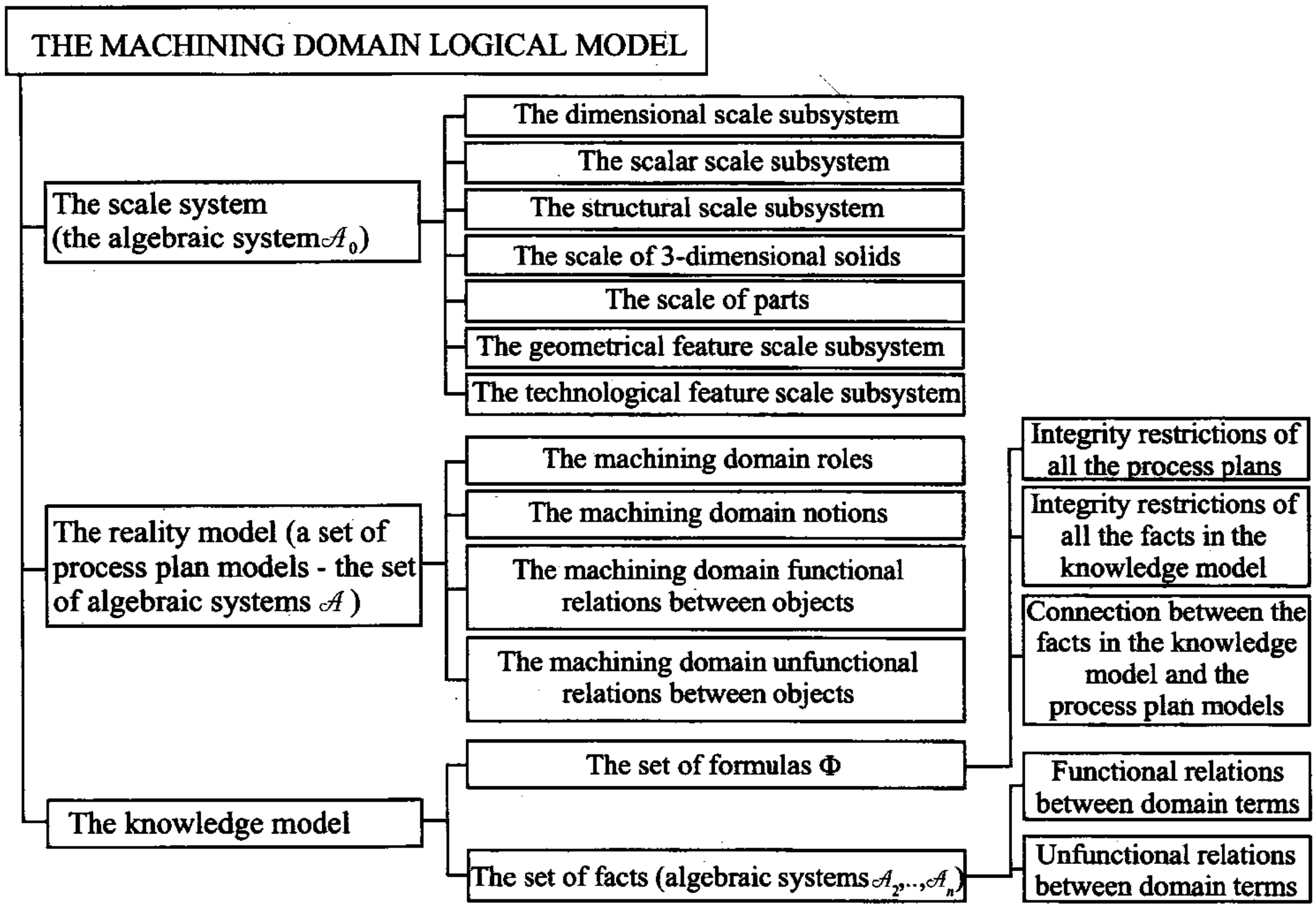
THE MACHINING DOMAIN LOGICAL MODEL
The scale system (the algebraic system $\mathcal{A}_0$)
The dimensional scale subsystem
The scalar scale subsystem
The structural scale subsystem
The scale of 3-dimensional solids
The scale of parts
The geometrical feature scale subsystem
The technological feature scale subsystem
The reality model (a set of process plan models - the set of algebraic systems $\mathcal{A}$)
The machining domain roles
The machining domain notions
The machining domain functional relations between objects
The machining domain unfunctional relations between objects
The knowledge model
The set of formulas Φ
The set of facts (algebraic systems $\mathcal{A}_2$,..,$\mathcal{A}_n$)
Integrity restrictions of all the process plans
Integrity restrictions of all the facts in the knowledge model
Connection between the facts in the knowledge model and the process plan models
Functional relations between domain terms
Unfunctional relations between domain terms

Figure 4.